# Learning to Order Things

**William W. Cohen**                    WCOHEN@RESEARCH.ATT.COM
**Robert E. Schapire**                  SCHAPIRE@RESEARCH.ATT.COM
**Yoram Singer**                        SINGER@RESEARCH.ATT.COM
*AT&T Labs, Shannon Laboratory, 180 Park Avenue*
*Florham Park, NJ 07932, USA*

## Abstract

There are many applications in which it is desirable to order rather than classify instances. Here we consider the problem of learning how to order instances given feedback in the form of preference judgments, *i.e.*, statements to the effect that one instance should be ranked ahead of another. We outline a two-stage approach in which one first learns by conventional means a binary *preference function* indicating whether it is advisable to rank one instance before another. Here we consider an on-line algorithm for learning preference functions that is based on Freund and Schapire's "Hedge" algorithm. In the second stage, new instances are ordered so as to maximize agreement with the learned preference function. We show that the problem of finding the ordering that agrees best with a learned preference function is NP-complete. Nevertheless, we describe simple greedy algorithms that are guaranteed to find a good approximation. Finally, we show how metasearch can be formulated as an ordering problem, and present experimental results on learning a combination of "search experts," each of which is a domain-specific query expansion strategy for a web search engine.

## 1. Introduction

Work in inductive learning has mostly concentrated on learning to classify. However, there are many applications in which it is desirable to order rather than classify instances. An example might be a personalized email filter that prioritizes unread mail. Here we will consider the problem of learning how to construct such orderings given feedback in the form of *preference judgments*, *i.e.*, statements that one instance should be ranked ahead of another.

Such orderings could be constructed based on a learned probabilistic classifier or regression model and in fact often are. For instance, it is common practice in information retrieval to rank documents according to their probability of relevance to a query, as estimated by a learned classifier for the concept "relevant document." An advantage of learning orderings directly is that preference judgments can be much easier to obtain than the labels required for classification learning.

For instance, in the email application mentioned above, one approach might be to rank messages according to their estimated probability of membership in the class of "urgent" messages, or by some numerical estimate of urgency obtained by regression. Suppose, however, that a user is presented with an ordered list of email messages, and elects to read the third message first. Given this election, it is not necessarily the case that message three is urgent, nor is there sufficient information to estimate any numerical urgency measures.

However, it seems quite reasonable to infer that message three should have been ranked ahead of the others. Thus, in this setting, obtaining preference information may be easier and more natural than obtaining the labels needed for a classification or regression approach.

Another application domain that requires ordering instances is *collaborative filtering*; see, for instance, the papers contained in Resnick and Varian (1997). In a typical collaborative filtering task, a user seeks recommendations, say, on movies that she is likely to enjoy. Such recommendations are usually expressed as ordered lists of recommended movies, produced by combining movie ratings supplied by other users. Notice that each user's movie ratings can be viewed as a set of preference judgements. In fact, interpreting ratings as preferences is advantageous in several ways: for instance, it is not necessary to assume that a rating of "7" means the same thing to every user.

In the remainder of this paper, we will investigate the following two-stage approach to learning how to order. In stage one, we learn a *preference function*, a two-argument function $\text{PREF}(u, v)$ which returns a numerical measure of how certain it is that $u$ should be ranked before $v$. In stage two, we use the learned preference function to order a set of new instances $X$; to accomplish this, we evaluate the learned function $\text{PREF}(u, v)$ on all pairs of instances $u, v \in X$, and choose an ordering of $X$ that agrees, as much as possible, with these pairwise preference judgments.

For stage one, we describe a specific algorithm for learning a preference function from a set of "ranking-experts". The algorithm is an on-line weight allocation algorithm, much like the weighted majority algorithm (Littlestone & Warmuth, 1994) and Winnow (Littlestone, 1988), and, more directly, Freund and Schapire's (1997) "Hedge" algorithm. For stage two, we show that finding a total order that agrees best with such a preference function is NP-complete. Nevertheless, we show that there are efficient greedy algorithms that always find a good approximation to the best ordering.

We then present some experimental results in which these algorithm are used to combine the results of several "search experts," each of which is a domain-specific query expansion strategy for a web search engine. Since our work touches several different fields we defer the discussion of related work to Sec. 6.

## 2. Preliminaries

Let $X$ be a set of instances. For simplicity, in this paper, we always assume that $X$ is finite. A *preference function* PREF is a binary function $\text{PREF} : X \times X \rightarrow [0, 1]$. A value of $\text{PREF}(u, v)$ which is close to 1 (respectively 0) is interpreted as a strong recommendation that $u$ should be ranked above (respectively, below) $v$. A value close to $1/2$ is interpreted as an abstention from making a recommendation. As noted earlier, the hypothesis of our learning system will be a preference function, and new instances will be ranked so as to agree as much as possible with the preferences predicted by this hypothesis.

In standard classification learning, a hypothesis is constructed by combining primitive features. Similarly, in this paper, a preference function will be a combination of primitive preference functions. In particular, we will typically assume the availability of a set of $N$ primitive preference functions $R_1, \ldots, R_N$. These can then be combined in the usual ways, for instance with a boolean or linear combination of their values. We will be especially interested in the latter combination method.

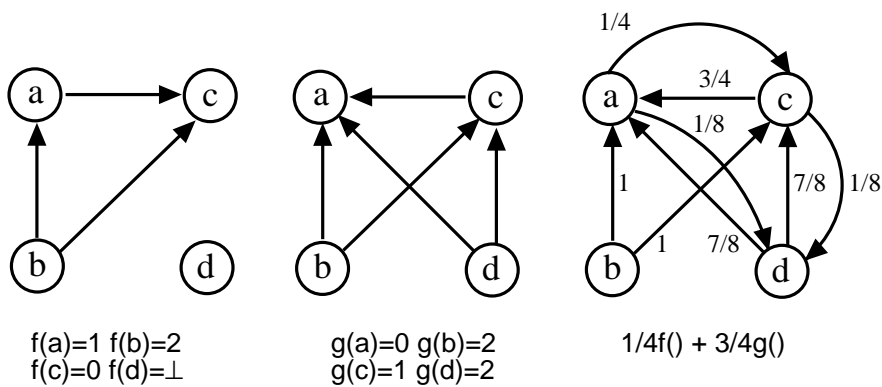

Figure 1: Left and middle: Two ordering functions and their graph representation. Right: The graph representation of the preference function created by a weighted ($\frac{1}{4}$ and $\frac{3}{4}$) combination of the two functions. Edges with weight of $\frac{1}{2}$ or 0 are omitted.

It is convenient to assume that the $R_i$'s are well-formed in certain ways. To this end, we introduce a special kind of preference function called a rank ordering which is defined by an ordering function. Let $S$ be a totally ordered set. We assume without loss of generality that $S \subseteq \mathbb{R}$. An *ordering function into $S$* is any function $f : X \to S$, where we interpret an inequality $f(u) > f(v)$ to mean that $u$ is ranked above $v$ by $f$. It is sometimes convenient to allow an ordering function to "abstain" and not give a preference for a pair $u$, $v$. We therefore allow $S$ to include a special symbol $\perp$ not in $\mathbb{R}$, and we interpret $f(u) = \perp$ to mean that $u$ is "unranked." We define the symbol $\perp$ to be incomparable to all the elements in $S$ (that is, $\perp \not\prec s$ and $s \not\prec \perp$ for all $s \in S$).

An ordering function $f$ induces the preference function $R_f$, defined as

$$R_f(u, v) = \begin{cases} 1 & \text{if } f(u) > f(v) \\ 0 & \text{if } f(u) < f(v) \\ \frac{1}{2} & \text{otherwise.} \end{cases}$$

We call $R_f$ a *rank ordering for $X$ into $S$*. If $R_f(u, v) = 1$, then we say that $u$ is preferred to $v$, or $u$ is ranked higher than $v$. Note that $R_f(u, v) = \frac{1}{2}$ if either $u$ or $v$ (or both) is unranked.

We will sometimes describe and manipulate preference functions as directed weighted graphs. The nodes of a graph correspond to the instances in $X$. Each pair $(u, v)$ is connected by a directed edge with weight $\mathrm{PREF}(u, v)$. Since an ordering function $f$ induces a preference function $R_f$, we can also describe ordering functions as graphs. In Fig. 1 we give an example of two ordering functions and their corresponding graphs. For brevity, we do not draw edges $(u, v)$ such that $\mathrm{PREF}(u, v) = \frac{1}{2}$ or $\mathrm{PREF}(u, v) = 0$.

To give a concrete example of rank orderings, imagine learning to order documents based on the words that they contain. To model this, let $X$ be the set of all documents in a repository, and for $N$ words $w_1, \ldots, w_N$, let $f_i(u)$ be the number of occurrences of word $w_i$ in document $u$. Then $R_{f_i}$ will prefer $u$ to $v$ whenever $w_i$ occurs more often in $u$ than $v$. As a second example, consider a metasearch application in which the goal is to combine the rankings of several web search engines on some fixed query. For $N$ search engines $e_1, \ldots, e_N$, one might define $f_i$ so that $R_{f_i}$ prefers web page $u$ to web page $v$ whenever $u$ is ranked ahead of $v$ in the list $L_i$ produced by the corresponding search engine. To do this, one could let $f_i(u) = -k$ for the web page $u$ appearing in the $k$-th position in the list $L_i$, and let $f_i(u) = -M$ (where $M > |L_i|$) for any web page $u$ not appearing in $L_i$.

Feedback from the user will be represented in a similar but more general way. We will assume that feedback is a set element pairs $(u, v)$, each representing an assertion of the form "$u$ should be preferred to $v$." This definition of feedback is less restricted than ordering functions. In particular, we will not assume that the feedback is consistent—cycles, such as $a > b > a$, will be allowed.

## 3. Learning a Combination of Ordering Functions

In this section, we consider the problem of learning a good linear combination of a set of ordering functions. Specifically, we assume access to a set of *ranking experts*, each of which generates an ordering function when provided with a set of instances. For instance, in a metasearch problem, each ranking expert might be a function that submits the user's query to a different search engine; the domain of instances might be the set of all web pages returned by any of the ranking experts; and the ordering function associated with each ranking expert might be represented as in the example above (*i.e.*, letting $f_i(u) = -k$ for the $k$-the web page $u$ returned by $i$-th search engine, and letting $f_i(u) = -M$ for any web page $u$ not retrieved by the $i$-th search engine). The user's feedback will be a set of pairwise preferences between web pages. This feedback may be obtained directly, for example, by asking the user to explicitly rank the URL's returned by the search engine; or the feedback may be obtained indirectly, for example, by measuring the time spent viewing each of the returned pages.

We note that for the metasearch problem, an approach that works directly with the numerical scores associated with the different search engines might not be feasible; these numerical scores might not be comparable across different search engines, or might not be provided by all search engines. Another problem is that most web pages will not be indexed by all search engines. This can be easily modeled in our setting: rather than letting $f_i(u) = -M$ for a web page $u$ that is not ranked by search engine $i$, one could let $f_i(u) = \bot$. This corresponds to the assumption that the search engine's preference for $u$ relative to ranked web pages is unknown.

We now describe a weight allocation algorithm that uses the preference functions $R_i$ to learn a preference function of the form $\text{PREF}(u, v) = \sum_{i=1}^{N} w_i R_i(u, v)$. We adopt the on-line learning framework first studied by Littlestone (1988) in which the weight $w_i$ assigned to each ranking expert $i$ is updated incrementally.

Formally, learning is assumed to take place in a sequence of rounds. On each round $t$, we assume the learning algorithm is provided with a set $X^t$ of instances to be ranked, for which each ranking expert $i \in \{1, \ldots, N\}$ provides an ordering function $f_i^t$. (In metasearch, for instance, $f_i^t$ is the ordering function associated with the list $L_i^t$ of web pages returned by the $i$-th ranking expert for the $t$-th query, and $X^t$ is the set of all web pages that appear in any of the lists $L_1^t, \ldots, L_N^t$.) Each ordering function $f_i^t$ induces a preference function $R_{f_i^t}$, which we denote for brevity by $R_i^t$. The learner may compute $R_i^t(u, v)$ for any and all preference functions $R_i^t$ and pairs $u, v \in X^t$ before producing a combined preference function $\text{PREF}^t$, which is then used to produce an ordering $\hat{\rho}^t$ of $X^t$. (Methods for producing an ordering from a preference function will be discussed below.)

After producing the ordering $\hat{\rho}^t$, the learner receives feedback from the environment. We assume that the feedback is an arbitrary set of assertions of the form "$u$ should be preferred to $v$." That is, the feedback on the $t$-th round is a set $F^t$ of pairs $(u, v)$.

The algorithm we propose for this problem is based on the "weighted majority algorithm" of Littlestone and Warmuth (1994) and, more directly, on Freund and Schapire's (1997) "Hedge" algorithm. We define the *loss* of a preference function $R$ with respect to the user's feedback $F$ as

$$\text{Loss}(R, F) = \frac{\sum_{(u,v) \in F} (1 - R(u, v))}{|F|} = 1 - \frac{1}{|F|} \sum_{(u,v) \in F} R(u, v) \ . \tag{1}$$

This loss has a natural probabilistic interpretation. If $R$ is viewed as a randomized prediction algorithm that predicts that $u$ will precede $v$ with probability $R(u, v)$, then $\text{Loss}(R, F)$ is the probability of $R$ disagreeing with the feedback on a pair $(u, v)$ chosen uniformly at random from $F$.

It is worth noting that the assumption on the form of the feedback can be further relaxed by allowing the user to indicate the degree to which she prefers $u$ over $v$. In this case, the loss should be normalized by the weighted sum of feedback pairs. Since this generalization is rather straightforward, we assume for brevity that the feedback is an unweighted set of assertions over element pairs.

We now can use the Hedge algorithm almost verbatim, as shown in Figure 2. The algorithm maintains a positive weight vector whose value at time $t$ is denoted by $\mathbf{w}^t = (w_1^t, \ldots, w_N^t)$. If there is no prior knowledge about the ranking experts, we set all initial weights to be equal so that $w_i^1 = 1/N$.

On each round $t$, the weight vector $\mathbf{w}^t$ is used to combine the preference functions of the different experts to obtain the preference function $\text{PREF}^t(u, v) = \sum_{i=1}^N w_i^t R_i^t(u, v)$. This preference function is next converted into an ordering $\hat{\rho}^t$ on the current set of elements $X^t$. For the purposes of this section, the method of producing an ordering is immaterial; in particular, any of the methods described in Sec. 4 could be used here. Based on this ordering, the user provides feedback $F^t$, and the loss for each preference function $\text{Loss}(R_i^t, F^t)$ is evaluated as in Eq. (1). Finally, the weight vector $\mathbf{w}^t$ is updated using the multiplicative rule

$$w_i^{t+1} = \frac{w_i^t \cdot \beta^{\text{Loss}(R_i^t, F^t)}}{Z_t}$$

where $\beta \in [0, 1]$ is a parameter, and $Z_t$ is a normalization constant, chosen so that the weights sum to one after the update. Thus, in each round, the weights of the ranking experts are adjusted so that experts producing preference functions with relatively large agreement with the feedback are increased.

We now give the theoretical rationale behind this algorithm. Freund and Schapire (1997) prove general results about Hedge which can be applied directly to this loss function. Their results imply almost immediately a bound on the cumulative loss of the preference function $\text{PREF}^t$ in terms of the loss of the best ranking expert, specifically:

**Allocate Weights for Ranking Experts**
**Parameters:** $\beta \in [0, 1]$, initial weight vector $\mathbf{w}^1 \in [0, 1]^N$ with $\sum_{i=1}^N w_i^1 = 1$
$\qquad\qquad N$ ranking experts, number of rounds $T$

**Do for** $t = 1, 2, \ldots, T$

1. Receive a set of elements $X^t$ and ordering functions $f_1^t, \ldots, f_N^t$. Let $R_i^t$ denote the preference function induced by $f_i^t$.

2. Compute a total order $\hat{\rho}^t$ which approximates

$$\text{PREF}^t(u, v) = \sum_{i=1}^N w_i^t R_i^t(u, v)$$

(Sec. 4 describes several ways of approximating a preference function with a total order.)

3. Order $X^t$ using $\hat{\rho}^t$.

4. Receive feedback $F^t$ from the user.

5. Evaluate losses $\text{Loss}(R_i^t, F^t)$ as defined in Eq. (1).

6. Set the new weight vector

$$w_i^{t+1} = \frac{w_i^t \cdot \beta^{\text{Loss}(R_i^t, F^t)}}{Z_t}$$

where $Z_t$ is a normalization constant, chosen so that $\sum_{i=1}^N w_i^{t+1} = 1$.

---

Figure 2: The on-line weight allocation algorithm.

**Theorem 1** *For the algorithm of Fig. 2,*

$$\sum_{t=1}^T \text{Loss}(\text{PREF}^t, F^t) \le a_\beta \min_i \sum_{t=1}^T \text{Loss}(R_i^t, F^t) + c_\beta \ln N$$

*where $a_\beta = \ln(1/\beta)/(1 - \beta)$ and $c_\beta = 1/(1 - \beta)$.*

Note that $\sum_t \text{Loss}(\text{PREF}^t, F^t)$ is the cumulative loss of the combined preference functions $\text{PREF}^t$, and $\sum_t \text{Loss}(R_i^t, F^t)$ is the cumulative loss of the $i$th ranking expert. Thus, Theorem 1 states that the cumulative loss of the combined preference functions will not be much worse than that of the best ranking expert.

**Proof:** We have that

$$\begin{aligned}
\text{Loss}(\text{PREF}^t, F^t) &= 1 - \frac{1}{F^t} \sum_{(u,v) \in F^t} \sum_i w_i^t R_i^t(u, v) \\
&= \sum_i w_i^t \left( 1 - \frac{1}{F^t} \sum_{(u,v) \in F^t} R_i^t(u, v) \right)
\end{aligned}$$

$$= \sum_i w_i^t \text{Loss}(R_i^t(u, v), F^t).$$

Therefore, by Freund and Schapire's (1997) Theorem 2,

$$\sum_{t=1}^T \text{Loss}(\text{PREF}^t, F^t) = \sum_{t=1}^T \sum_i w_i^t \text{Loss}(R_i^t(u, v), F^t)$$

$$\leq a_\beta \min_i \sum_{t=1}^T \text{Loss}(R_i^t, F^t) + c_\beta \ln N.$$

$\square$

Of course, we are not interested in the loss of $\text{PREF}^t$ (since it is not an ordering), but rather in the performance of the actual ordering $\hat{\rho}^t$ computed by the learning algorithm. Fortunately, the losses of these can be related using a kind of triangle inequality. Let

$$\text{DISAGREE}(\rho, \text{PREF}) = \sum_{u, v: \rho(u) > \rho(v)} (1 - \text{PREF}(u, v)) . \qquad (2)$$

**Theorem 2** *For any* $\text{PREF}$*,* $F$ *and total order defined by an ordering function* $\rho$*,*

$$\text{Loss}(R_\rho, F) \leq \frac{\text{DISAGREE}(\rho, \text{PREF})}{|F|} + \text{Loss}(\text{PREF}, F). \qquad (3)$$

**Proof:** For $x, y \in [0, 1]$, let us define $d(x, y) = x(1 - y) + y(1 - x)$. We now show that $d$ satisfies the triangle inequality. Let $x$, $y$ and $z$ be in $[0, 1]$, and let $X$, $Y$ and $Z$ be independent Bernoulli ($\{0, 1\}$-valued) random variables with probability of outcome 1 equal to $x$, $y$ and $z$, respectively. Then

$$d(x, z) = \Pr[X \neq Z]$$
$$= \Pr[(X \neq Y \wedge Y = Z) \vee (X = Y \wedge Y \neq Z)]$$
$$\leq \Pr[X \neq Y \vee Y \neq Z]$$
$$\leq \Pr[X \neq Y] + \Pr[Y \neq Z]$$
$$= d(x, y) + d(y, z).$$

For $[0, 1]$-valued functions $f, g$ defined on $X \times X$, we next define

$$D(f, g) = \sum_{u, v: u \neq v} d(f(u, v), g(u, v)).$$

Clearly, $D$ also satisfies the triangle inequality.

Let $\chi_F$ be the characteristic function of $F$ so that $\chi_F : X \times X \to \{0, 1\}$ and $\chi_F(u, v) = 1$ if and only if $(u, v) \in F$. Then from the definition of Loss and DISAGREE, we have

$$|F| \text{Loss}(R_\rho, F) = D(R_\rho, \chi_F)$$
$$\leq D(R_\rho, \text{PREF}) + D(\text{PREF}, \chi_F)$$
$$= \text{DISAGREE}(\rho, \text{PREF}) + |F| \text{Loss}(\text{PREF}, F).$$

$\square$

Notice that the learning algorithm Hedge minimizes the second term on the right hand side of Eq. (3). Below, we consider the problem of finding an ordering $\rho$ which minimizes the first term, namely, DISAGREE.

## 4. Ordering Instances with a Preference Function

### 4.1 Measuring the Quality of an Ordering

We now consider the complexity of finding a total order that agrees best with a learned preference function. To analyze this, we must first quantify the notion of agreement between a preference function PREF and an ordering. One natural notion is the following: Let $X$ be a set, PREF be a preference function, and let $\rho$ be a total ordering of $X$, expressed again as an ordering function (*i.e.*, $\rho(u) > \rho(v)$ if and only if $u$ is above $v$ in the order). For the analysis of this section, it is convenient to use the measure $\text{AGREE}(\rho, \text{PREF})$, which is defined to be the sum of $\text{PREF}(u, v)$ over all pairs $u, v$ such that $u$ is ranked above $v$ by $\rho$:

$$\text{AGREE}(\rho, \text{PREF}) = \sum_{u,v : \rho(u) > \rho(v)} \text{PREF}(u, v). \qquad (4)$$

Clearly, AGREE is a linear transformation of the measure DISAGREE introduced in Eq. (2), and hence maximizing AGREE is equivalent to minimizing DISAGREE. This definition is also closely related to similarity metrics used in decision theory and information processing (Kemeny & Snell, 1962; Fishburn, 1970; Roberts, 1979; French, 1989; Yao, 1995) (see the discussion in Sec. 6).

### 4.2 Finding an Optimal Ordering is Hard

Ideally one would like to find a $\rho$ that maximizes $\text{AGREE}(\rho, \text{PREF})$. The general optimization problem is of little interest in our setting, since there are many constraints on the preference function that are imposed by the learning algorithm. Using the learning algorithm of Sec. 3, for instance, PREF will always be a linear combination of simpler functions. However, the theorem below shows that this optimization problem is NP-complete even if PREF is restricted to be a linear combination of well-behaved preference functions. In particular, the problem is NP-complete even if all the primitive preference functions used in the linear combination are rank orderings which map into a set $S$ with only three elements, one of which may or may not be $\perp$. (Clearly, if $S$ consists of more than three elements then the problem is still hard.)

**Theorem 3** *The following decision problem is NP-complete for any set $S$ with $|S| \geq 3$:*
    Input: *A rational number $\kappa$; a set $X$; a collection of $N$ ordering functions $f_i : X \to S$; and a preference function* PREF *defined as*

$$\text{PREF}(u, v) = \sum_{i=1}^{N} w_i R_{f_i}(u, v) \qquad (5)$$

*where $\mathbf{w} = (w_1, \ldots, w_N)$ is a rational weight vector in $[0, 1]^N$ with $\sum_{i=1}^{N} w_i = 1$.*
    Question: *Does there exist a total order $\rho$ such that $\text{AGREE}(\rho, \text{PREF}) \geq \kappa$?*

**Proof:** The problem is clearly in NP since a nondeterministic algorithm can guess a total order and check the weighted number of agreements in polynomial time.

To prove that the problem is NP-hard we reduce from CYCLIC-ORDERING (Galil & Megido, 1977; Gary & Johnson, 1979), defined as follows: "Given a set $A$ and a collection $C$ of ordered triples $(a, b, c)$ of distinct elements from $A$, is there a one-to-one function $f : A \rightarrow \{1, 2, \ldots, |A|\}$ such that for each $(a, b, c) \in C$ we have either $f(a) > f(b) > f(c)$ or $f(b) > f(c) > f(a)$ or $f(c) > f(a) > f(b)$?"

Without loss of generality, $S$ is either $\{0, 1, \perp\}$ or $\{0, 1, 2\}$. We first show that the problem of finding an optimal total order is hard when $S = \{0, 1, \perp\}$. Given an instance of CYCLIC-ORDERING, we let $X = A$. For each triplet $t = (a, b, c)$ we will introduce three ordering functions $f_{t,1}$, $f_{t,2}$, and $f_{t,3}$, and define them so that $f_{t,1}(a) > f_{t,1}(b)$, $f_{t,2}(b) > f_{t,2}(c)$, and $f_{t,3}(c) > f_{t,3}(a)$. To do this, we let $f_{t,1}(a) = f_{t,2}(b) = f_{t,3}(c) = 1$, $f_{t,1}(b) = f_{t,2}(c) = f_{t,3}(a) = 0$, and $f_{t,i}(\cdot) = \perp$ in all other cases. We let the weight vector be uniform, so that $w_{t,i} = \frac{1}{3|C|}$. Let

$$\kappa = \frac{5}{3} + \frac{|A|(|A|-1)/2 - 3}{2}.$$

Define $R_t(u, v) = \sum_{i=1}^{3} w_{t,i} R_{f_{t,i}}(u, v)$, which is the contribution of these three functions to $\mathrm{PREF}(u, v)$. Notice that for any triplet $t = (a, b, c) \in C$, $R_t(a, b) = \frac{2}{3|C|}$ whereas $R_t(b, a) = \frac{1}{3|C|}$, and similarly for $b, c$ and $c, a$. In addition, for any pair $u, v \in A$ such that at least one of them does not appear in $t$, we get that $R_t(u, v) = \frac{1}{2|C|}$. Since a total order $\rho$ can satisfy at most two of the three conditions $\rho(a) > \rho(b)$, $\rho(b) > \rho(c)$, and $\rho(c) > \rho(a)$, the largest possible weighted number of agreements associated with this triple is exactly $\kappa/|C|$.

If the number of weighted agreements is at least $\kappa$, it must be exactly $\kappa$, by the argument above; and if there are exactly $\kappa$ weighted agreements, then the total order must satisfy exactly 2 out of the possible 3 relations for each three elements that form a triplet from $C$. Thus, the constructed rank ordering instance will be positive if and only if the original CYCLIC-ORDERING instance is positive.

The case for $S = \{0, 1, 2\}$ uses a similar construction; however, for each triplet $t = (a, b, c)$, we define six ordering functions, $f_{t,1}^j$, $f_{t,2}^j$, and $f_{t,3}^j$, where $j \in \{0, 1\}$. The basic idea here is to replace each $f_{t,i}$ with two functions, $f_{t,i}^0$ and $f_{t,i}^1$, that agree on the single ordering constraint associated with $f_{t,i}$, but disagree on all other orderings. For instance, we will define these functions so that $f_{t,1}^j(a) > f_{t,1}^j(b)$ for $j = 0$ and $j = 1$, but for all other pairs $u, v$, $f_{t,1}^1(u) > f_{t,1}^1(v)$ iff $f_{t,1}^0(v) > f_{t,1}^0(u)$. Averaging the two orderings $f_{t,1}^0$ and $f_{t,1}^1$ will thus yield the same preference expressed by the original function $f_{t,1}$ (*i.e.*, a preference for $a > b$ only).

In more detail, we let $f_{t,1}^j(a) = f_{t,2}^j(b) = f_{t,3}^j(c) = 2 - j$, $f_{t,1}^j(b) = f_{t,2}^j(c) = f_{t,3}^j(a) = 1 - j$, and $f_{t,i}^j(\cdot) = 2j$ in all other cases. We again let the weight vector be uniform, so that $w_{t,i}^j = \frac{1}{6|C|}$. Similar to the first case, we define $R_t(u, v) = \sum_{i,j} w_{t,i} R_{f_{t,i}^j}(u, v)$. It can be verified that $R_t$ is identical to the $R_t$ constructed in the first case. Therefore, by the same argument, the constructed rank ordering instance will be positive if and only if the original CYCLIC-ORDERING instance is positive. $\square$

Although this problem is hard when $|S| \geq 3$, the next theorem shows that it becomes tractable for linear combinations of rank orderings into a set $S$ of size two. Of course, when $|S| = 2$, the rank orderings are really only binary classifiers. The fact that this special case is tractable underscores the fact that manipulating orderings (even relatively simple ones) can be computationally more difficult than performing the corresponding operations on binary classifiers.

**Theorem 4** *The following optimization problem is solvable in linear time:*
   **Input:** *A set $X$; a set $S$ with $|S| = 2$; a collection of $N$ ordering functions $f_i : X \to S$; and a preference function PREF defined by Eq. (5).*
   **Output:** *A total order defined by an ordering function $\rho$ which maximizes* AGREE$(\rho, \text{PREF})$.

**Proof:**   Assume without loss of generality that the two-element set $S$ is $\{0, 1\}$, and define $\rho(u) = \sum_i w_i f_i(u)$. We now show that any total order[1] consistent with $\rho$ maximizes AGREE$(\rho, \text{PREF})$. Fix a pair $u, v \in X$ and let

$$q_{b_1 b_2} = \sum_{i \text{ s.t. } f_i(u) = b_1, f_i(v) = b_2} w_i \ .$$

We can now rewrite $\rho$ and PREF as

$$\begin{array}{rclcrcl}
\rho(u) & = & q_{10} + q_{11} & \qquad & \text{PREF}(u, v) & = & q_{10} + \tfrac{1}{2} q_{11} + \tfrac{1}{2} q_{00} \\
\rho(v) & = & q_{01} + q_{11} & \qquad & \text{PREF}(v, u) & = & q_{01} + \tfrac{1}{2} q_{11} + \tfrac{1}{2} q_{00} \ .
\end{array}$$

Note that both $\rho(u) - \rho(v)$ and $\text{PREF}(u, v) - \text{PREF}(v, u)$ are equal to $q_{10} - q_{01}$. Hence, if $\rho(u) > \rho(v)$ then $\text{PREF}(u, v) > \text{PREF}(v, u)$. Therefore, for each pair $u, v \in X$, the order defined by $\rho$ agrees on *all* pairs with the pairwise preference defined by PREF. In other words, we have shown that

$$\text{AGREE}(\rho, \text{PREF}) = \sum_{\{u, v\}} \max\{\text{PREF}(u, v), \text{PREF}(v, u)\} \tag{6}$$

where the sum is over all unordered pairs. Clearly, the right hand side of Eq. (6) maximizes the right hand side of Eq. (4) since at most one of $(u, v)$ or $(v, u)$ can be included in the latter sum.   □

### 4.3 Finding an Approximately Optimal Ordering

Theorem 3 implies that we are unlikely to find an efficient algorithm that finds the optimal total order for a weighted combination of rank orderings. Fortunately, there do exist efficient algorithms for finding an *approximately* optimal total order. In fact, finding a good total order is closely related to the problem of finding the minimum feedback arc set, for which there exist good approximation algorithms; see, for instance, (Shmoys, 1997) and the references therein. However, the algorithms that achieve the good approximation results for the minimum feedback arc set problem are based on (or further approximate) a linear-programming relaxation (Seymour, 1995; Even, Naor, Rao, & Schieber, 1996; Berger & Shor, 1997; Even, Naor, Schieber, & Sudan, 1998) which is rather complex to implement and quite slow in practice.

---

1. Notice that in case of a tie, so that $\rho(u) = \rho(v)$ for distinct $u, v$, $\rho$ defines only a partial order. The theorem holds for any total order which is consistent with this partial order, *i.e.*, for any $\rho'$ so that $\rho(u) > \rho(v) \Rightarrow \rho'(u) > \rho'(v)$.

**Algorithm Greedy-Order**
**Inputs:** an instance set $X$; a preference function PREF
**Output:** an approximately optimal ordering function $\hat{\rho}$
**let** $V = X$
**for** each $v \in V$ **do** $\pi(v) = \sum_{u \in V} \text{PREF}(v, u) - \sum_{u \in V} \text{PREF}(u, v)$
**while** $V$ is non-empty **do**
    **let** $t = \arg\max_{u \in V} \pi(u)$
    **let** $\hat{\rho}(t) = |V|$
    $V = V - \{t\}$
    **for** each $v \in V$ **do** $\pi(v) = \pi(v) + \text{PREF}(t, v) - \text{PREF}(v, t)$
**endwhile**

Figure 3: The greedy ordering algorithm.

We describe instead a simple greedy algorithm which is very simple to implement. Figure 3 summarizes the greedy algorithm. As we will shortly demonstrate, this algorithm produces a good approximation to the best total order.

The algorithm is easiest to describe by thinking of PREF as a directed weighted graph, where initially, the set of vertices $V$ is equal to the set of instances $X$, and each edge $u \rightarrow v$ has weight $\text{PREF}(u, v)$. We assign to each vertex $v \in V$ a *potential* value $\pi(v)$, which is the weighted sum of the outgoing edges *minus* the weighted sum of the ingoing edges. That is,

$$\pi(v) = \sum_{u \in V} \text{PREF}(v, u) - \sum_{u \in V} \text{PREF}(u, v) .$$

The greedy algorithm then picks some node $t$ that has maximum potential[2], and assigns it a rank by setting $\hat{\rho}(t) = |V|$, effectively ordering it ahead of all the remaining nodes. This node, together with all incident edges, is then deleted from the graph, and the potential values $\pi$ of the remaining vertices are updated appropriately. This process is repeated until the graph is empty. Notice that nodes removed in subsequent iterations will have progressively smaller and smaller ranks.

As an example, consider the preference function defined by the leftmost graph of Fig. 4. (This graph is identical to the weighted combination of the two ordering functions from Fig. 1.) The initial potentials the algorithm assigns are: $\pi(b) = 2$, $\pi(d) = 3/2$, $\pi(c) = -5/4$, and $\pi(a) = -9/4$. Hence, $b$ has maximal potential. It is given a rank of 4, and then node $b$ and all incident edges are removed from the graph.

The result is the middle graph of Fig. 4. After deleting $b$, the potentials of the remaining nodes are updated: $\pi(d) = 3/2$, $\pi(c) = -1/4$, and $\pi(a) = -5/4$. Thus, $d$ will be assigned rank $|V| = 3$ and removed from the graph, resulting in the rightmost graph of Fig. 4.

After updating potentials again, $\pi(c) = 1/2$ and $\pi(a) = -1/2$. Now $c$ will be assigned rank $|V| = 2$ and removed, resulting in a graph containing the single node $a$, which will finally be assigned the rank $|V| = 1$. The ordering produced by the greedy algorithm is thus $b > d > c > a$.

---

2. Ties can be broken arbitrarily in case of two or more nodes with the same potential.

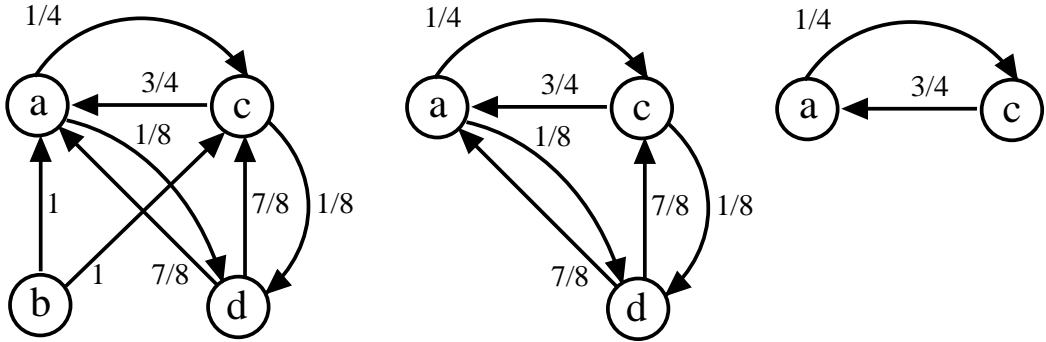

Figure 4: Behavior of the greedy ordering algorithm. The leftmost graph is the original input. From this graph, node $b$ will be assigned maximal rank and deleted, leading to the middle graph; from this graph, node $d$ will deleted, leading to the rightmost graph. In the rightmost graph, node $c$ will be ranked ahead of node $a$, leading the total ordering $b > d > c > a$.

The next theorem shows that this greedy algorithm comes within a factor of two of optimal.

**Theorem 5** *Let* OPT(PREF) *be the weighted agreement achieved by an optimal total order for the preference function* PREF, *and let* APPROX(PREF) *be the weighted agreement achieved by the greedy algorithm. Then*

$$\text{APPROX}(\text{PREF}) \geq \frac{1}{2}\text{OPT}(\text{PREF}) \ .$$

**Proof:** Consider the edges that are incident on the node $v_j$ which is selected on the $j$-th repetition of the **while** loop of Figure 3. The ordering produced by the algorithm will agree with all of the outgoing edges of $v_j$ and disagree with all of the ingoing edges. Let $a_j$ be the sum of the weights of the outgoing edges of $v_j$, and $d_j$ be the sum of the weights of the ingoing edges. Clearly APPROX(PREF) $\geq \sum_{j=1}^{|V|} a_j$. However, at every repetition, the total weight of all incoming edges must equal the total weight of all outgoing edges. This means that $\sum_{v \in V} \pi(v) = 0$, and hence for the node $v^\star$ that has maximal potential, $\pi(v^\star) \geq 0$. Thus on every repetition $j$, it must be that $a_j \geq d_j$, so we have that

$$\text{OPT}(\text{PREF}) \ \leq \ \sum_{j=1}^{|V|}(a_j + d_j) \ \leq \ \sum_{j=1}^{|V|}(a_j + a_j) \ \leq \ 2 \cdot \text{APPROX}(\text{PREF}).$$

The first inequality holds because OPT(PREF) can at best include every edge in the graph, and since every edge is removed exactly once, each edge must contribute to some $a_j$ or some $d_j$. $\quad\square$

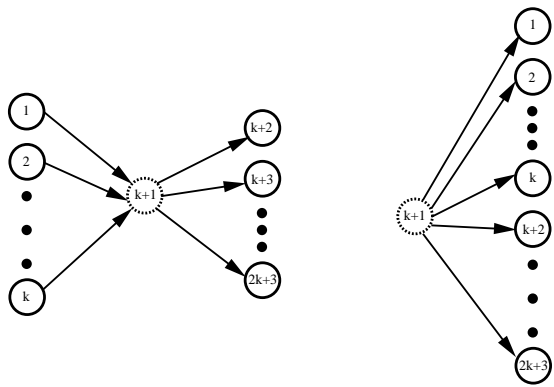

Figure 5: An example of a graph (left) for which the node-based greedy algorithm achieves an approximation factor of $\frac{1}{2}$ by constructing the partial order on the right.

In passing, we note that there are other natural greedy algorithms that do not achieve good approximations. Consider, for example, an algorithm that starts from a graph consisting of all the nodes but with no edges, and iteratively adds the highest weighted edge in the graph, while avoiding cycles. It can be shown that this algorithm can produce a very poor partial order, given an adversarially chosen graph; there are cases where the optimal total order achieves a multiplicative factor of $O(|V|)$ more weighted agreements than this "edge-based" greedy algorithm.

### 4.4 Improvements to the Greedy Algorithm

The approximation factor of two given in Theorem 5 is tight. That is, there exist problems for which the greedy algorithm approximation is worse than the optimal solution by a factor arbitrarily close to two. Consider the graph shown on the left-hand side of Fig. 5. An optimal total order ranks the instances according to their position in the figure, left to right, breaking ties randomly, and achieves OPT(PREF) $= 2k+2$ weighted agreements. However, the greedy algorithm picks the node labeled $k + 1$ first and orders all the remaining nodes randomly, achieving as few as APPROX(PREF) $= k + 2$ agreements. For large $k$, the ratio APPROX(PREF)/OPT(PREF) approaches $\frac{1}{2}$.

For graph of Figure 5, there is another simple algorithm which produces an optimal ordering: since the graph is already a partial order, picking any total order consistent with this partial order gives an optimal result. To cope with problems such as the one of Figure 5, we devised an improvement to the greedy algorithm which combines a greedy method with topological sorting. The aim of the improvement is to find better approximations for graphs which are composed of many strongly connected components.

As before, the modified algorithm is easiest to describe by thinking of PREF as a weighted directed graph. Recall that for each pair of nodes $u$ and $v$, there exist two edges: one from $u$ to $v$ with weight PREF$(u, v)$ and one from $v$ to $u$ with weight PREF$(v, u)$. In the modified greedy algorithm we will pre-process the graph. For each pair of nodes, we

**Algorithm SCC-Greedy-Order**
**Inputs:** an instance set $X$; a preference function PREF
**Output:** an approximately optimal ordering function $\hat{\rho}$

**Define** $\text{PREF}'(u, v) = \max\{\text{PREF}(u, v) - \text{PREF}(v, u),\ 0\}$ .
**Find** strongly connected components $U_1, \ldots, U_k$ of the graph $G = (V, E)$ where

$$V = X \text{ and } E = \{(u, v) \mid \text{PREF}'(u, v) > 0\} \ .$$

**Order** the strongly connected components in any way consistent with the partial order $<_{\text{scc}}$:

$$U <_{\text{scc}} U' \text{ iff } \exists u \in U, u' \in U' : (u, u') \in E$$

**Use** algorithm **Greedy-Order** or full enumeration to order the instances within each component $U_i$ according to $\text{PREF}'$.

---

Figure 6: The improved greedy ordering algorithm.

remove the edge with the smaller weight and set the weight of the other edge to be

$$\mid \text{PREF}(v, u) - \text{PREF}(u, v) \mid \ .$$

For the special case where $\text{PREF}(v, u) = \text{PREF}(u, v) = \frac{1}{2}$, we remove both edges. In the reduced graph, there is at most one directed edge between each pair of nodes. Note that the greedy algorithm would behave identically on the transformed graph since it is based on the weighted *differences* between the incoming and outgoing edges.

We next find the strongly connected components[3] of the reduced graph, ignoring (for now) the weights. One can now split the edges of the reduced graph into two classes: *inter-component* edges connect nodes $u$ and $v$, where $u$ and $v$ are in different strongly connected components; and *intra-component* edges connect nodes $u$ and $v$ from the same strongly connected component. It is straightforward to verify that any optimal order agrees with all the inter-component edges. Put another way, if there is an edge from node $u$ to node $v$ of two different connected components in the reduced graph, then $\rho(u) > \rho(v)$ for any optimal total order $\rho$.

The first step of the improved algorithm is thus to totally order the strongly connected components in some way consistent with the partial order defined by the inter-component edges. More precisely, we pick a total ordering for the components consistent with the partial order $<_{\text{scc}}$, defined as follows: for components $U$ and $U'$, $U <_{\text{scc}} U'$ iff there is an edge from some node $u \in U$ to some node $u' \in U'$ in the reduced graph.

We next order the nodes within each strongly connected component, thus providing a total order of all nodes. Here the greedy algorithm can be used. As an alternative, in cases where a component contains only a few elements (say at most five), one can find the optimal order between the elements of the component by a brute-force approach, *i.e.*, by full enumeration of all permutations.

---

3. Two nodes $u$ and $v$ are in the same strongly connected component iff there are directed paths from $u$ to $v$ and from $v$ to $u$.

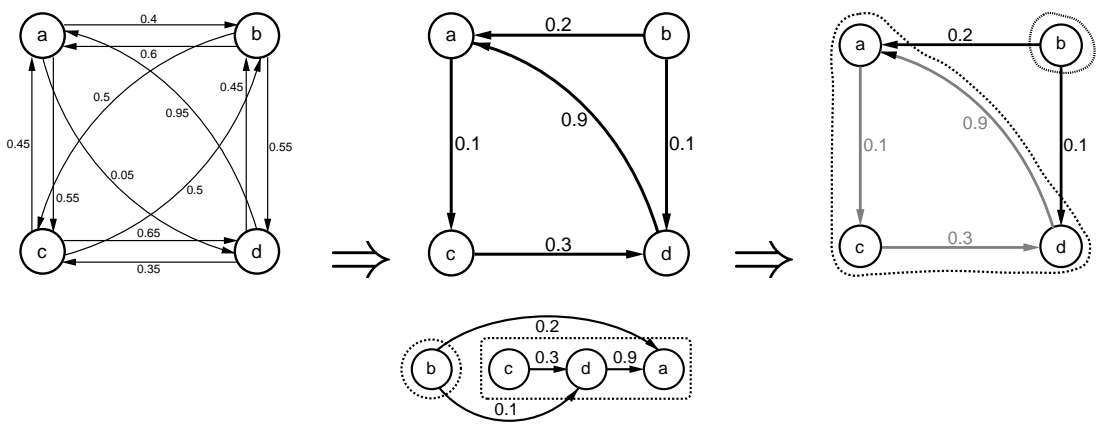

Figure 7: An illustration of the approximation algorithm for finding a total order from a weighted combination of ordering functions. The original graph (top left) is reduced by removing at least one edge for each edge-pair $(u, v)$ and $(v, u)$ (middle). The strongly connected components are then found (right). Finally, an ordering is found within each strongly connected component which yield the order $b > c > d > a$ (bottom).

The improved algorithm is summarized in Figure 6 and illustrated in Figure 7. There are four elements in Figure 7 which constitute two strongly connected components in the reduced graph ($\{b\}$ and $\{a, c, d\}$). Therefore, $b$ is assigned the top rank and ranked above $a$, $c$ and $d$. If the brute-force algorithm were used to order the components, then we would check all 3! permutations between $a$, $c$ and $d$ and output the total order $b > c > d > a$, which is the optimal order in this toy example.

In the worst case, the reduced graph contains only a single strongly connected component. In this case, the improved algorithm generates the same ordering as the greedy algorithm. However, in the experiments on metasearch problems described in Sec. 5, many of the strongly connected components are small; the average size of a strongly connected component is less than five. In cases such as these, the improved algorithm will often improve on the simple greedy algorithm.

## 4.5 Experiments with the Ordering Algorithms

Ideally, each algorithm would be evaluated by determining how closely it approximates the optimal ordering on large, realistic problems. Unfortunately, finding the optimal ordering for large graphs is impractical. We thus performed two sets of experiments with the ordering algorithms described above. In the first set of experiments, we evaluated the algorithms on small graphs—specifically, graphs for which the optimal ordering could be feasibly found with brute-force enumeration. In these experiments, we measure the "goodness" of the resulting orderings relative to the optimal ordering. In the second set of experiments, we evaluated the algorithms on large graphs for which the optimal orderings are unknown. In these experiments, we compute a "goodness" measure which depends on the total weight of all edges, rather than the optimal ordering.

In addition to the simple greedy algorithm and its improvement, we also considered the following simple randomized algorithm: pick a permutation at random, and then output the better of that permutation and its reverse. It can be easily shown that this algorithm achieves the same approximation bound on expected performance as the greedy algorithm. (Briefly, one of the two permutations must agree with at least half of the weighted edges in the graph.) The random algorithm can be improved by repeating the process, *i.e.*, examining many random permutations and their reverses, and choosing the permutation that achieves the largest number of weighted agreements.

In a first set of experiments, we compared the performance of the greedy approximation algorithm, the improved algorithm which first finds strongly connected components, and the randomized algorithm on graphs of nine or fewer elements. For each number of elements, we generated 10,000 random graphs by choosing $\text{PREF}(u, v)$ uniformly at random, and setting $\text{PREF}(v, u)$ to $1 - \text{PREF}(u, v)$. For the randomized algorithm, we evaluated $10n$ random permutations (and their reverses) where $n$ is the number of instances (nodes). To have a fair comparison between the different algorithms on the smaller graphs, we always used the greedy algorithm (rather than a brute-force algorithm) to order the elements of each strongly connected component of a graph.

To evaluate the algorithms, we examined the reduced graph and calculated the average ratio of the weights of the edges chosen by the approximation algorithm to the weights of the edges that were chosen by the optimal order. More precisely, let $\rho$ be the optimal order and $\hat{\rho}$ be an order chosen by an approximation algorithm. Then for each random graph, we calculated

$$\frac{\displaystyle\sum_{u, v \,:\, \hat{\rho}(u) > \hat{\rho}(v)} \max\{\text{PREF}(u, v) - \text{PREF}(v, u), 0\}}{\displaystyle\sum_{u, v \,:\, \rho(u) > \rho(v)} \max\{\text{PREF}(u, v) - \text{PREF}(v, u), 0\}} \, .$$

If this measure is 0.9, for instance, then the total weight of the edges in the total order picked by the approximation algorithm is 90% of the corresponding figure for the optimal algorithm.

We averaged the above ratios over all random graphs of the same size. The results are shown on the left hand side of Figure 8. On the right hand side of the figure, we show the average running time for each of the algorithms as a function of the number of elements. When the number of ranked elements is more than five, the greedy algorithms outperform the randomized algorithm, while their running time is much smaller. Thus, if a full enumeration had been used to find the optimal order of small strongly connected components, the approximation would have been consistently better than the randomized algorithm.

We note that the greedy algorithm also generally performs better on average than the lower bound given in Theorem 5. In fact, combining the greedy algorithm with pre-partitioning of the graph into strongly connected components often yields the optimal order.

In the second set of experiments, we measured performance and running time for larger random graphs. Since for large graphs we cannot find the optimal solution by brute-force enumeration, we use as a "goodness" measure the ratio of the weights of the edges that were left in the reduced graph after applying an approximation algorithm to the *total* weight of

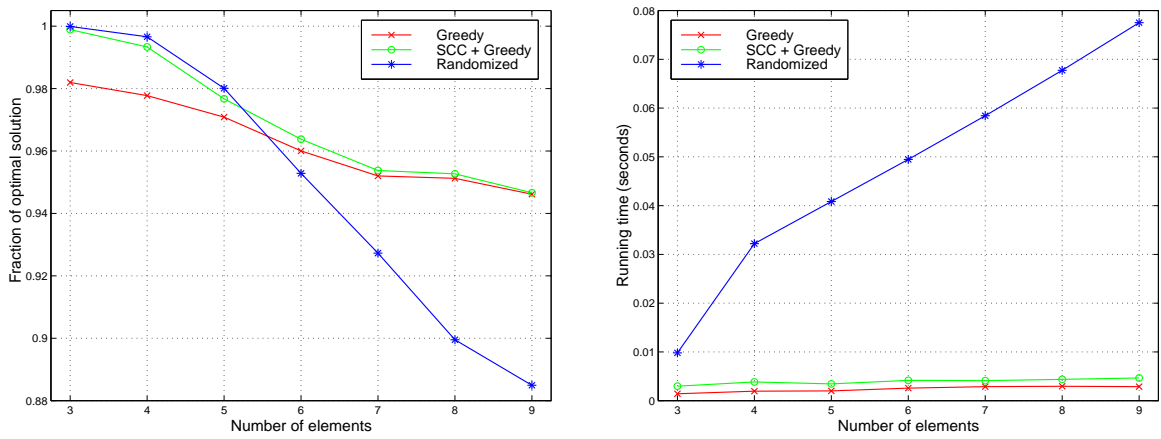

Figure 8: Comparison of goodness (left) and the running time (right) of the approximations achieved by the greedy algorithms and the randomized algorithm as a function of the number of ranked elements for random preference functions with 3 through 9 elements.

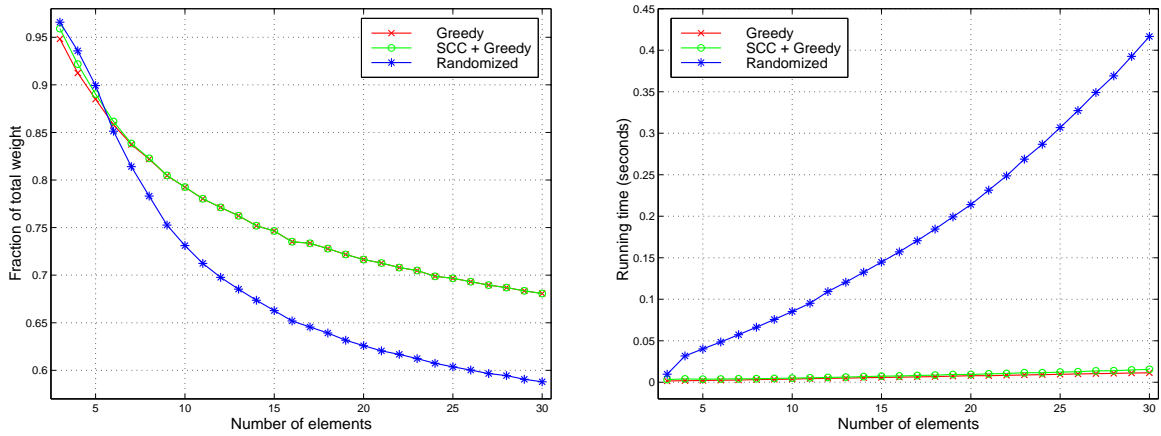

Figure 9: Comparison of goodness (left) and the running time (right) of the approximations achieved by the greedy algorithms and the randomized algorithm as a function of the number of ranked elements for random preference functions with 3 through 30 elements. Note that the graphs for Greedy and SCC+Greedy coincide for most of the points.

edges in the graph. That is, for each random graph we calculated

$$\frac{\displaystyle\sum_{u,\,v\,:\,\hat{\rho}(u)\,>\,\hat{\rho}(v)} \max\{\mathrm{PREF}(u,v) - \mathrm{PREF}(v,u), 0\}}{\displaystyle\sum_{u,\,v} \max\{\mathrm{PREF}(u,v) - \mathrm{PREF}(v,u), 0\}} \; .$$

We ran the three algorithms with the same parameters as above (*i.e.*, 10,000 random graphs). The results are given in Figure 9. The advantage of the greedy algorithms over the randomized algorithm is even more apparent on these larger problems. Note also that for large graphs the performance of the two greedy algorithms is indistinguishable. This is mainly due to the fact that large *random* graphs are strongly connected with high probability.

To summarize the experiments, when there are six or more elements the greedy algorithm clearly outperforms the randomized algorithm even if many randomly chosen permutations are examined. Furthermore, the improved algorithm which first finds the strongly connected components outperforms the randomized algorithm for all graph sizes. In practice the improved greedy algorithm achieves very good approximations—within about 5 percent of optimal, for the cases in which optimal graphs can be feasibly found.

## 5. Experimental Results for Metasearch

So far, we have described a method for learning a preference function, and a means of converting a preference function into an ordering of new instances. We will now present some experimental results in learning to order. In particular, we will describe results on learning to combine the orderings of several web "search experts" using the algorithm of Figure 2 to learn a preference function, and the simple greedy algorithm to order instances using the learned preference function. The goals of these experiments are to illustrate the type of problems that can be solved with our method; to empirically evaluate the learning method; to evaluate the ordering algorithm on large, non-random graphs, such as might arise in a realistic application; and to confirm the theoretical results of the preceding sections. We thus restrict ourselves to comparing the learned orderings to individual search experts, as is suggested by Theorem 1, rather than attempt to compare this application of learning-to-order with previous experimental techniques for metasearch, e.g., (Lochbaum & Streeter, 1989; Kantor, 1994; Boyan, Freitag, & Joachims, 1994; Bartell, Cottrell, & Belew, 1994).

We note that this metasearch problem exhibits several properties that suggest a general approach such as ours. For instance, approaches that learn to combine similarity scores are not applicable, since the similarity scores of web search engines are often unavailable. In the experiments presented here, the learning algorithm was provided with ordered lists for each search engine without any associated scores. To further demonstrate the merits of our approach, we also describe experiments with partial feedback—that is, with preference judgments that are less informative than the relevance judgments more typically used in improving search engines.

| ML Search Experts | UNIV Search Experts |
|---|---|
| NAME | NAME |
| "NAME" | "NAME" |
| title:"NAME" | "NAME" PLACE |
| NAME +LASTNAME title:"home page" | title:NAME |
| NAME +LASTNAME title:homepage | title:"NAME" |
| NAME +LASTNAME machine learning | title:"NAME" PLACE |
| NAME +LASTNAME "machine learning" | NAME title:"home page" |
| NAME +LASTNAME case based reasoning | NAME title:"homepage" |
| NAME +LASTNAME "case based reasoning" | NAME welcome |
| NAME +LASTNAME PLACE | NAME url:index.html |
| NAME +LASTNAME "PLACE" | NAME url:home.html |
| NAME +LASTNAME url:index.html | "NAME" title:"home page" |
| NAME +LASTNAME url:home.html | "NAME" title:"homepage" |
| NAME +LASTNAME url:~*LASTNAME* | "NAME" welcome |
| NAME +LASTNAME url:~LASTNAME | "NAME" url:index.html |
| NAME +LASTNAME url:LASTNAME | "NAME" url:home.html |
| | "NAME" PLACE title:"home page" |
| | "NAME" PLACE title:"homepage" |
| | "NAME" PLACE welcome |
| | "NAME" PLACE url:index.html |
| | "NAME" PLACE url:home.html |

Table 1: Search (and ranking) experts used in the metasearch experiments. In the associated queries, NAME is replaced with the person's (or university's) full name, LASTNAME with the person's last name, and PLACE is replaced with the person's affiliation (or university's location). Sequences of words enclosed in quotes must appear as a phrase, and terms prefixed by `title:` and `url:` must appear in that part of the web page. Words prefixed by a "+" must appear in the web page; other words may or may not appear.

## 5.1 Test Problems and Encoding

We chose to simulate the problem of learning a domain-specific search engine—*i.e.*, an engine that searches for pages of a particular, narrow type. Ahoy! (Shakes, Langheinrich, & Etzioni, 1997) is one instance of such a domain-specific search engine. As test cases, we picked two problems: retrieving the home pages of machine learning researchers (ML), and retrieving the home pages of universities (UNIV). To obtain sample queries, we obtained a listing of machine learning researchers, identified by name and affiliated institution, together with their home pages,[4] and a similar list for universities, identified by name and (sometimes) geographical location.[5] Each entry on a list was viewed as a query, with the associated URL the sole relevant web page.

---

4. From http://www.aic.nrl.navy.mil/~aha/research/machine-learning.html, a list maintained by David Aha.

5. From Yahoo!

We then constructed a series of special-purpose "search experts" for each domain. These were implemented as query expansion methods which converted a name/affiliation pair (or a name/location pair) to a likely-seeming Altavista query. For example, one expert for the UNIV domain searched for the university name appearing as a phrase, together with the phrase "home page" in the title; another expert for the ML domain searched for all the words in the person's name plus the words "machine" and "learning," and further enforces a strict requirement that the person's last name appear. Overall, we defined 16 search experts for the ML domain and 22 for the UNIV domain; these are summarized in Table 1. Each search expert returned the top 30 ranked web pages. In the ML domain, there were 210 searches for which at least one search expert returned the named home page; for the UNIV domain, there were 290 such searches. The task of the learning system is to find an appropriate way of combining the output of these search experts.

To give a more precise description of the search experts, for each query $t$, we first constructed the set $X^t$ consisting of all web pages returned by all of the expanded queries defined by the search experts. Next, each search expert $i$ was represented as a preference function $R_i^t$. We chose these preference functions to be rank orderings defined with respect to an ordering function $f_i^t$ in the natural way: we assigned a rank of $f_i^t = 30$ to the first listed page, $f_i^t = 29$ to the second-listed page, and so on, finally assigning a rank of $f_i^t = 0$ to every page not retrieved in the top 30 by the expanded query associated with expert $i$.

To encode feedback, we considered two schemes. In the first, we simulated complete relevance feedback—that is, for each query, we constructed feedback in which the sole relevant page was preferred to all other pages. In the second, we simulated the sort of feedback that could be collected from "click data"—*i.e.*, from observing a user's interactions with a metasearch system. For each query, after presenting a ranked list of pages, we noted the rank of the one relevant web page. We then constructed a feedback ranking in which the relevant page is preferred to all preceding pages. This would correspond to observing which link the user actually followed, and making the assumption that this link was preferred to previous links.

It should be emphasized that both of these forms of feedback are simulated, and contain less noise than would be expected from real user data. In reality some fraction of the relevance feedback would be missing or erroneous, and some fraction of click data would not satisfy the assumption stated above.

## 5.2 Evaluation and Results

To evaluate the expected performance of a fully-trained system on novel queries in this domain, we employed leave-one-out testing. For each query $t$, we trained the learning system on all the other queries, and then recorded the rank of the learned system on query $t$. For complete relevance feedback, this rank is invariant of the ordering of the training examples, but for the "click data" feedback, it is not; the feedback collected at each stage depends on the behavior of the partially learned system, which in turn depends on the previous training examples. Thus for click data training, we trained on 100 randomly chosen permutations of the training data and recorded the median rank for $t$.

### 5.2.1 Performance Relative to Individual Experts

The theoretical results provide a guarantee of performance relative to the performance of the best individual search (ranking) expert. It is therefore natural to consider comparing the performance of the learned system to the best of the individual experts. However, for each search expert, only the top 30 ranked web pages for a query are known; if the single relevant page for a query is not among these top 30, then it is impossible to compute any natural measures of performance for this query. This complicates any comparison of the learned system to the individual search experts.

However, in spite of the incomplete information about the performance of the search experts, it is usually possible to tell if the learned system ranks a web page higher than a particular expert.[6] Motivated by this, we performed a sign test: we compared the rank of the learning systems to the rank given by each search expert, checking to see whether this rank was lower, and discarding queries for which this comparison was impossible. We then used a normal approximation to the binomial distribution to test the following two null hypotheses (where the probability is taken over the distribution from which the queries are drawn):

**H1.** With probability at least 0.5, the search expert performs better than the learning system (*i.e.*, gives a lower rank to the relevant page than the learning system does.)

**H2.** With probability at least 0.5, the search expert performs no worse than the learning system (*i.e.*, gives an equal or lower rank to the relevant page.)

In training, we explored learning rates in the range $[0.001, 0.999]$. For complete feedback in the ML domain, hypothesis H1 can be rejected with high confidence ($p > 0.999$) for every search expert and every learning rate $0.01 \leq \beta \leq 0.99$. The same holds in the UNIV domain for all learning rates $0.02 \leq \beta \leq 0.99$. The results for click data training are nearly as strong, except that 2 of the 22 search experts in the UNIV domain show a greater sensitivity to the learning rate: for these engines, H1 can only be rejected with high confidence for $0.3 \leq \beta \leq 0.6$. To summarize, with high confidence, in both domains, the learned ranking system is no worse than any individual search expert for moderate values of $\beta$.

Hypothesis H2 is more stringent since it can be rejected only if we are sure that the learned system is strictly *better* than the expert. With complete feedback in the ML domain and $0.3 \leq \beta \leq 0.8$, hypothesis H2 can be rejected with confidence $p > 0.999$ for 14 of the 16 search experts. For the remaining two experts the learned system does perform better more often, but the difference is not significant. In the UNIV domain, the results are similar. For $0.2 \leq \beta \leq 0.99$, hypothesis H2 can be rejected with confidence $p > 0.999$ for 21 of the 22 search experts, and the learned engine tends to perform better than the single remaining expert.

Again, the results for click data training are only slightly weaker. In the ML domain, hypothesis H2 can be rejected for all but three experts for all but the most extreme learning rates; in the UNIV domain, hypothesis H2 can be rejected for all but two experts for $0.4 \leq \beta \leq 0.6$. For the remaining experts and learning rates the differences are not statistically

---

6. The only time this cannot be determined is when neither the learned system nor the expert ranks the relevant web pages in the top 30, a case of little practical interest.

significant; however, it is not always the case that the learned engine tends to perform better.

To summarize the experiments, for moderate values of $\beta$ the learned system is, with high confidence, strictly better than most of the search experts in both domains, and never significantly worse than any expert. When trained with full relevance judgments, the learned system performs better on average than any individual expert.

### 5.2.2 OTHER PERFORMANCE MEASURES

We measured the number of queries for which the correct web page was in the top $k$ ranked pages, for various values of $k$. These results are shown in Figure 10. Here the lines show the performance of the learned systems (with $\beta = 0.5$, a generally favorable learning rate) and the points correspond to the individual experts. In most cases, the learned system closely tracks the performance of the best expert at every value of $k$. This is especially interesting since no single expert is best at all values of $k$.

The final graph in this figure investigates the sensitivity of this measure to the learning rate $\beta$. As a representative illustration, we varied $\beta$ in the ML domain and plotted the top-$k$ performance of the system learned from complete feedback for three values of $k$. Note that performance is roughly comparable over a wide range of values for $\beta$.

Another plausible measure of performance is the average rank of the (single) relevant web page. We computed an approximation to average rank by artificially assigning a rank of 31 to every page that was either unranked, or ranked above rank 30. (The latter case is to be fair to the learned system, which is the only one for which a rank greater than 30 is possible.) A summary of these results for $\beta = 0.5$ is given in Table 2, together with some additional data on top-$k$ performance. In the table, we give the top-$k$ performance for three values of $k$, and average rank for several ranking systems: the two learned systems; the naive query, *i.e.*, the person or university's name; and the single search expert that performed best with respect to each performance measure. Note that not all of these experts are distinct since several experts scored the best on more than one measure.

The table illustrates the robustness of the learned systems, which are nearly always competitive with the best expert for every performance measure listed. The only exception to this is that the system trained on click data trails the best expert in top-$k$ performance for small values of $k$. It is also worth noting that in both domains, the naive query (simply the person or university's name) is not very effective: even with the weaker click data feedback, the learned system achieves a 36% decrease in average rank over the naive query in the ML domain, and a 46% decrease in the UNIV domain.

To summarize the experiments, on these domains the learned system not only performs much better than naive search strategies, but also consistently performs at least as well as, and perhaps slightly better than, any single domain-specific search expert. This observation holds regardless of the performance metric considered; for nearly every metric we computed, the learned system always equals, and usually exceeds, the performance of the search expert that is best for that metric. Finally, the performance of the learned system is almost as good with the weaker "click data" training as with complete relevance feedback.

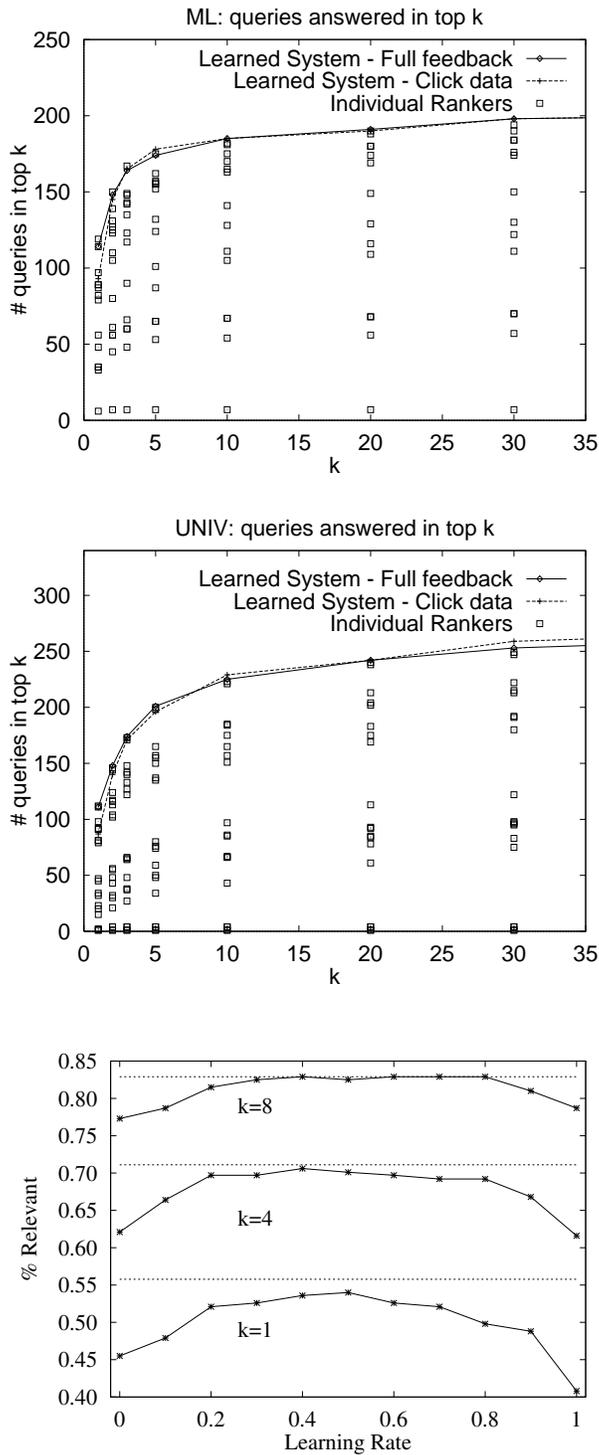

Figure 10: Top and middle: Performance of the learned system versus individual experts for two different domains. Bottom: the percentage of time the relevant web page was in the top-$k$ list for $k = 1,4$, and 8.

| | ML Domain | | | | University Domain | | | |
|---|---|---|---|---|---|---|---|---|
| | Top 1 | Top 10 | Top 30 | Avg Rank | Top 1 | Top 10 | Top 30 | Avg Rank |
| Learned (Full Feed.) | 114 | **185** | **198** | **4.9** | 111 | 225 | 253 | **7.8** |
| Learned (Click Data) | 93 | **185** | **198** | **4.9** | 87 | **229** | **259** | **7.8** |
| Naive | 89 | 165 | 176 | 7.7 | 79 | 157 | 191 | 14.4 |
| Best (Top 1) | **119** | 170 | 184 | 6.7 | **112** | 221 | 247 | 8.2 |
| Best (Top 10) | 114 | 182 | 190 | 5.3 | 111 | 223 | 249 | 8.0 |
| Best (Top 30) | 97 | 181 | 194 | 5.6 | 111 | 223 | 249 | 8.0 |
| Best (Avg Rank) | 114 | 182 | 190 | 5.3 | 111 | 223 | 249 | 8.0 |

Table 2: Comparison of learned systems and individual search queries.

## 6. Related Work

Problems that involve ordering and ranking have been investigated in various fields such as decision theory, the social sciences, information retrieval and mathematical economics (Black, 1958; Kemeny & Snell, 1962; Cooper, 1968; Fishburn, 1970; Roberts, 1979; Salton & McGill, 1983; French, 1989; Yao, 1995). Among the wealth of literature on the subject, the closest to ours appears to be the work of Kemeny and Snell (1962) which was extended by Yao (1995) and used by Balabanovíc and Shoham (1997) in their FAB collaborative filtering system. These works use a similar notion of ordering functions and feedback; however, they assume that both the ordering functions and the feedback are complete and transitive. Hence, it is not possible to leave elements unranked, or to have inconsistent feedback which violates the transitivity requirements. It is therefore difficult to combine and fuse inconsistent and incomplete orderings in the Kemeny and Snell model.

There are also several related intractability results. Most of them are concerned with the difficulty in reaching consensus in voting systems based on preference ordering. Specifically, Bartholdi, Tovey and Trick (1989) study the problem of finding a winner in an election when the preferences of all voters are irreflexive, antisymmetric, transitive, and complete. Thus, their setting is more restrictive than ours. They study two similar schemes to decide on a winner of an election. The first was invented by Dodgson (1876) (better known by his pen name, Lewis Carroll) and the second is due to Kemeny (1959). For both models, they show that the problem of finding a winner in an election is NP-hard. Among these two models, the one suggested by Kemeny is the closest to ours. However, as mentioned above, this model is more restrictive as it does not allow voters to abstain (preferences are required to be complete) or to be inconsistent (all preferences are transitive).

As illustrated by the experiments, the problem of learning to rank is closely related to the problem of combining the results of different search engines. Many methods for this have been proposed by the information retrieval community, and many of these are adaptive, using relevance judgments to make an appropriate choice of parameters. However, generally, rankings are combined by combining the scores that were used to rank documents (Lochbaum & Streeter, 1989; Kantor, 1994). It is also frequently assumed that other properties of the objects (documents) to be ranked are available, such as word frequencies. In contrast, in our experiments, instances are atomic entities with no associated properties except for their position in various rank-orderings. Similarly, we make minimal assump-tions about the rank-orderings—in particular, we do not assume scores are available. Our methods are thus applicable to a broader class of ranking problems.

General optimization methods have also been adopted to adjust parameters of an IR system so as to improve agreement with a set of user-given preference judgments. For instance, Boyan, Freitag, and Joachims (1994) use simulated annealing to improve agreement with "click data," and Bartell, Cottrell and Belew (1994) use conjugate gradient descent to choose parameters for a linear combination of scoring functions, each associated with a different search expert. Typically, such approaches offer few guarantees of efficiency, optimality, or generalization performance.

Another related task is *collection fusion*. Here, several searches are executed on disjoint subsets of a large collection, and the results are combined. Several approaches to this problem that do not rely on combining ranking scores have been described (Towell, Voorhees, Gupta, & Johnson-Laird, 1995; Voorhees, Gupta, & Johnson-Laird, 1994). However, although the problem is superficially similar to the one presented here, the assumption that the different search engines index *disjoint* sets of documents actually makes the problem quite different. In particular, since it is impossible for two engines to give different relative orderings to the same pair of documents, combining the rankings can be done relatively easily.

Etzioni et al. (1996) formally considered another aspect of metasearch—the task of optimally combining information sources with associated costs and time delays. Our formal results are disjoint from theirs, as they assume that every query has a single recognizable correct answer, rendering ordering issues unimportant.

There are many other applications in machine learning, reinforcement learning, neural networks, and collaborative filtering that employ ranking and preferences, e.g., (Utgoff & Saxena, 1987; Utgoff & Clouse, 1991; Caruana, Baluja, & Mitchell, 1996; Resnick & Varian, 1997), While our work is not directly relevant, it might be possible to use the framework suggested in this paper in similar settings. This is one of our future research goals.

Finally, we would like to note that the framework and algorithms presented in this paper can be extended in several ways. Our current research focuses on efficient batch algorithms for combining preference functions, and on using restricted ranking experts for which the problem of finding an optimal total ordering can be solved in polyomial time (Freund, Iyer, Schapire, & Singer, 1998).

## 7. Conclusions

In many applications, it is desirable to order rather than classify instances. We investigated a two-stage approach to learning to order in which one first learns a preference function by conventional means, and then orders a new set of instances by finding the total ordering that best approximates the preference function. The preference function that is learned is a binary function $PREF(u, v)$, which returns a measure of confidence reflecting how likely it is that $u$ is preferred to $v$. This is learned from a set of "experts" which suggest specific orderings, and from user feedback in the form of assertions of the form "$u$ should be preferred to $v$".

We have presented two sets of results on this problem. First, we presented an online learning algorithm for learning a weighted combination of ranking experts which is based on an adaptation of Freund and Schapire's Hedge algorithm. Second, we explored the complexity of the problem of finding a total ordering that agrees best with a preference function. We showed that this problem is NP-complete even in a highly restrictive case, namely, preference predicates that are linear combinations of a certain class of well-behaved "experts" called rank orderings. However, we also showed that for any preference predicate, there is a greedy algorithm that always obtains a total ordering that is within a factor of two of optimal. We also presented an algorithm that first divides the set of instances into strongly connected components and then uses the greedy algorithm (or full enumeration, for small components) to find an approximately good order within large strongly connected components. We found that this approximation algorithm works very well in practice and often finds the best order.

We also presented experimental results in which these algorithms were used to combine the results of a number of "search experts," each of which corresponds to a domain-specific strategy for searching the web. We showed that in two domains, the learned system closely tracks and often exceeds the performance of the best of these search experts. These results hold for either traditional relevance feedback models of learning, or from weaker feedback in the form of simulated "click data." The performance of the learned systems also clearly exceeds the performance of more naive approaches to searching.

## Acknowledgments

We would like to thank Noga Alon, Edith Cohen, Dana Ron, and Rick Vohra for numerous helpful discussions. An extended abstract of this paper appeared in *Advances in Neural Information Processing Systems 10*, MIT Press, 1998.

## References

Balabanovíc, M., & Shoham, Y. (1997). FAB: Content-based, collaborative recommendation. *Communications of the ACM, 40*(3), 66–72.

Bartell, B., Cottrell, G., & Belew, R. (1994). Automatic combination of multiple ranked retrieval systems. In *Seventeenth Annual International ACM SIGIR Conference on Research and Development in Information Retrieval*.

Bartholdi, J., Tovey, C., & Trick, M. (1989). Voting schemes for which it can be difficult to tell who won the elections. *Social Choice and Welfare, 6*, 157–165.

Berger, B., & Shor, P. (1997). Tight bounds for the acyclic subgraph problem. *Journal of Algorithms, 25*, 1–18.

Black, D. (1958). *Theory of Committees and Elections*. Cambridge University Press.

Boyan, J., Freitag, D., & Joachims, T. (1994). A machine learning architecture for optimizing web search engines. Tech. rep. WS-96-05, American Association of Artificial Intelligence.

Caruana, R., Baluja, S., & Mitchell, T. (1996). Using the future to 'Sort Out' the present: Rankprop and multitask learning for medical risk evaluation. In *Advances in Neural Information Processing Systems (NIPS) 8*.

Cooper, W. (1968). Expected search length: A single measure of retrieval effectiveness based on the weak ordering action of retrieval systems. *American Documentation*, *19*, 30–41.

Dodgson, C. (1876). *A method for taking votes on more than two issues*. Clarendon Press, Oxford. Reprinted with discussion in (Black, 1958).

Etzioni, O., Hanks, S., Jiang, T., Karp, R. M., Madani, O., & Waarts, O. (1996). Efficient information gathering on the internet. In *Proceedings of the 37th Annual Symposium on Foundations of Computer Science (FOCS-96)* Burlington, Vermont. IEEE Computer Society Press.

Even, G., Naor, J., Rao, S., & Schieber, B. (1996). Divide-and-conquer approximation algorithms via spreading metrics. In *36th Annual Symposium on Foundations of Computer Science (FOCS-96)*, pp. 62–71 Burlington, Vermont. IEEE Computer Society Press.

Even, G., Naor, J., Schieber, B., & Sudan, M. (1998). Approximating minimum feedback sets and multicuts in directed graphs. *Algorithmica*, *20*(2), 151–174.

Fishburn, F. (1970). *Utility Theory for Decision Making*. Wiley, New York.

French, S. (1989). *Decision Theory: An Introduction to the Mathematics of Rationality*. Ellis Horwood Series in Mathematics and Its Applications.

Freund, Y., Iyer, R., Schapire, R., & Singer, Y. (1998). An efficient boosting algorithm for combining preferences. In *Machine Learning: Proceedings of the Fifteenth International Conference*.

Freund, Y., & Schapire, R. (1997). A decision-theoretic generalization of on-line learning and an application to boosting. *Journal of Computer and System Sciences*, *55*(1), 119–139.

Galil, Z., & Megido, N. (1977). Cyclic ordering is NP-complete. *Theoretical Computer Science*, *5*, 179–182.

Gary, M., & Johnson, D. (1979). *Computers and Intractability: A Guide to the Theory of NP-completeness*. W. H. Freeman and Company, New York.

Kantor, P. (1994). Decision level data fusion for routing of documents in the TREC3 context: a best case analysis of worst case results. In *Proceedings of the third text retrieval conference (TREC-3)*.

Kemeny, J. (1959). Mathematics without numbers. *Daedalus*, *88*, 571–591.

Kemeny, J., & Snell, J. (1962). *Mathematical Models in the Social Sciences*. Blaisdell, New York.

Littlestone, N. (1988). Learning quickly when irrelevant attributes abound: A new linear-threshold algorithm. *Machine Learning*, *2*(4).

Littlestone, N., & Warmuth, M. (1994). The weighted majority algorithm. *Information and Computation*, *108*(2), 212–261.

Lochbaum, K., & Streeter, L. (1989). Comparing and combining the effectiveness of latent semantic indexing and the ordinary vector space model for information retrieval. *Information processing and management*, *25*(6), 665–676.

Resnick, P., & Varian, H. (1997). Introduction to special section on Recommender Systems. *Communication of the ACM*, *40*(3).

Roberts, F. (1979). *Measurement theory with applications to decision making, utility, and social sciences*. Addison Wesley, Reading, MA.

Salton, G., & McGill, M. (1983). *Introduction to Modern Information Retrieval*. McGraw-Hill.

Seymour, P. (1995). Packing directed circuits fractionally. *Combinatorica*, *15*, 281–288.

Shakes, J., Langheinrich, M., & Etzioni, O. (1997). Dynamic reference sifting: a case study in the homepage domain. In *Proceedings of WWW6*.

Shmoys, D. (1997). Cut problems and their application to divide-and-conquer. In Hochbaum, D. (Ed.), *Approximation algorithms for NP-Hard Problems*. PWS Publishing Company, New York.

Towell, G., Voorhees, E., Gupta, N., & Johnson-Laird, B. (1995). Learning collection fusion strategies for information retrieval. In *Machine Learning: Proceedings of the Twelfth International Conference* Lake Taho, California. Morgan Kaufmann.

Utgoff, P., & Clouse, J. (1991). Two kinds of training information for evaluation function learning. In *Proceedings of the Ninth National Conference on Artificial Intelligence (AAAI-91)*, pp. 596–600 Cambridge, MA. AAAI Press/MIT PRess.

Utgoff, P., & Saxena, S. (1987). Learning a preference predicate. In *Proceedings of the Fourth International Workshop on Machine Learning*, pp. 115–121 San Francisco, CA. Morgan Kaufmann.

Voorhees, E., Gupta, N., & Johnson-Laird, B. (1994). The collection fusion problem. In *Seventeenth Annual International ACM SIGIR Conference on Research and Development in Information Retrieval*.

Yao, Y. (1995). Measuring retrieval effectiveness based on user preference of documents. *Journal of the American Society for Information Science*, *46*(2), 133–145.